\documentclass{article}

\usepackage{microtype}
\usepackage{graphicx}
\usepackage{subfigure}
\usepackage{booktabs} %

\usepackage{hyperref}
\usepackage{epsfig}
\usepackage{enumitem}
\usepackage{amssymb}

\usepackage{amsmath,amsfonts,bm}

\def\eqref#1{equation~\ref{#1}}

\def\1{\bm{1}}

\def\rvu{{\mathbf{i}}}

\def\rvu{{\mathbf{u}}}
\def\rvv{{\mathbf{v}}}

\def\ervu{{\textnormal{u}}}
\def\ervv{{\textnormal{v}}}

\def\rmP{{\mathbf{P}}}

\DeclareMathAlphabet{\mathsfit}{\encodingdefault}{\sfdefault}{m}{sl}
\SetMathAlphabet{\mathsfit}{bold}{\encodingdefault}{\sfdefault}{bx}{n}

\def\gG{{\mathcal{G}}}

\def\sD{{\mathbb{D}}}
\def\sD{{\mathbb{D}}}

\def\sR{{\mathbb{R}}}

\def\sV{{\mathbb{V}}}

\def\sX{{\mathbb{X}}}
\def\sY{{\mathbb{Y}}}

\newcommand{\R}{\mathbb{R}}

\newcommand{\Amat}{{\bf A}}

\newcommand{\Cmat}{{\bf C}}

\newcommand{\Qmat}{{\bf Q}}

\newcommand{\Tmat}{{\bf T}}

\newcommand{\Xmat}{{\bf X}}
\newcommand{\Ymat}{{\bf Y}}

\newcommand{\av}{{\boldsymbol a}}

\newcommand{\bv}{{\boldsymbol b}}

\newcommand{\lv}{{\boldsymbol l}}

\newcommand{\pv}{{\boldsymbol p}}
\newcommand{\qv}{{\boldsymbol q}}

\newcommand{\xv}{{\boldsymbol x}}
\newcommand{\yv}{{\boldsymbol y}}

\newcommand{\gammav}{{\boldsymbol \gamma}}

\newcommand{\thetav}{{\boldsymbol \theta}}

\newcommand{\muv}{{\boldsymbol \mu}}
\newcommand{\nuv}{{\boldsymbol \nu}}

\newcommand{\Lcal}{\mathcal{L}}

\usepackage{hyperref}
\usepackage{url}
\usepackage{adjustbox}
\usepackage{wrapfig}
\usepackage{amsthm}
\usepackage{adjustbox}
\theoremstyle{definition}
\newtheorem{definition}{Definition}[section]

\newcommand{\ie}{\textit{i.e.}}

\usepackage[accepted]{icml2020}

\begin{document}

\twocolumn[
\icmltitle{Graph Optimal Transport for Cross-Domain Alignment}

\icmlsetsymbol{equal}{*}

\begin{icmlauthorlist}
\icmlauthor{Liqun Chen}{duke}
\icmlauthor{Zhe Gan}{micro}
\icmlauthor{Yu Cheng}{micro}
\icmlauthor{Linjie Li}{micro}
\icmlauthor{Lawrence Carin}{duke}
\icmlauthor{Jingjing Liu}{micro}

\end{icmlauthorlist}

\icmlaffiliation{duke}{Duke University}
\icmlaffiliation{micro}{Microsoft Dynamics 365 AI Research}
\icmlcorrespondingauthor{Liqun Chen}{liqun.chen@duke.edu}
\icmlcorrespondingauthor{Zhe Gan}{zhe.gan@microsoft.com}

\icmlkeywords{cross-domain alignment, deep learning, optimal transport}

\vskip 0.3in
]

\printAffiliationsAndNotice{Most of this work was done when the first author was
an intern at Microsoft.}  %

\begin{abstract}
Cross-domain alignment between two sets of entities
(\emph{e.g.}, objects in an image, words in a sentence) is fundamental to both computer vision and natural language processing. Existing methods mainly focus on designing advanced attention mechanisms to simulate soft alignment, with no training signals to \emph{explicitly} encourage alignment. The learned attention matrices are also dense and lacks interpretability. We propose Graph Optimal Transport (GOT), a principled framework that germinates from recent advances in Optimal Transport (OT). In GOT, cross-domain alignment is formulated as a graph matching problem, by representing entities into a dynamically-constructed graph. Two types of OT distances are considered: ($i$) Wasserstein distance (WD) for node (entity) matching; and ($ii$) Gromov-Wasserstein distance (GWD) for edge (structure) matching. 
Both WD and GWD can be incorporated into existing neural network models, effectively acting as a drop-in regularizer.  
The inferred transport plan also yields \emph{sparse} and \emph{self-normalized} alignment, enhancing the interpretability of the learned model. Experiments show consistent outperformance of GOT over baselines across a wide range of tasks, including image-text retrieval, visual question answering, image captioning, machine translation, and text summarization. 
\end{abstract}

\section{Introduction}
\label{intro}
Cross-domain Alignment (CDA), which aims to associate related entities across different domains,
plays a central role in a wide range of deep learning tasks, such as 
image-text retrieval~\cite{karpathy2015deep,SCAN}, 
visual question answering (VQA)~\cite{malinowski2014multi,antol2015vqa}, and machine translation~\citep{bahdanau2014neural,vaswani2017attention}. Considering VQA as an example, in order to understand the contexts in the image and the question, a model needs to interpret the latent alignment between regions in the input image and words in the question.
Specifically, a good model should:
($i$) identify entities of interest in both the image (\emph{e.g.}, objects/regions) and the question (\emph{e.g.}, words/phrases),
($ii$) quantify both intra-domain (within the image or sentence) and cross-domain relations between these entities, and then 
($iii$) design good metrics for measuring the quality of cross-domain alignment drawn from these relations, in order to optimize towards better results.

CDA is particularly challenging as it constitutes a \emph{weakly supervised learning} task. That is, only paired spaces of entity are given (\emph{e.g.}, an image paired with a question), while the ground-truth relations between these entities are not provided (\emph{e.g.}, no supervision signal for a ``dog'' region in an image aligning with the word ``dog'' in the question). 
State-of-the-art methods principally focus on designing advanced attention mechanisms to simulate soft alignment~\cite{bahdanau2014neural,xu2015show,yang2016hierarchical,yang2016stacked,vaswani2017attention}. For example, \citet{SCAN, kim2018bilinear, yu2019deep} have shown that learned co-attention can model dense interactions between entities  and infer cross-domain latent alignments for vision-and-language tasks. 
Graph attention has also been applied to relational reasoning for image captioning~\cite{yao2018exploring} and VQA~\cite{li2019relation}, such as graph attention network (GAT)~\cite{velivckovic2017graph} for capturing relations between entities in a graph via masked attention, and graph matching network (GMN)~\cite{li2019graph} for graph alignment via cross-graph soft attention. 
However, conventional attention mechanisms are guided by task-specific losses, with no training signal to \emph{explicitly} encourage alignment. And the learned attention matrices are often dense and uninterpretable, thus inducing less effective relational inference.  

We address whether there is a more principled approach to scalable discovery of cross-domain relations. To explore this, we present Graph Optimal Transport (GOT),\footnote{Another GOT framework was proposed in~\citet{maretic2019got} for graph comparison. We use the same acronym for the proposed algorithm; however, our method is very different from theirs.} a new framework for cross-domain alignment that leverages recent advances in Optimal Transport (OT). OT-based learning aims to optimize for distribution matching via minimizing the cost of transporting one distribution to another. We extend this to CDA (here a domain can be language, images, videos, etc.). The transport plan is thus redefined as transporting the distribution of embeddings from one domain (\emph{e.g.}, language) to another (\emph{e.g.}, images). By minimizing the cost of the learned transport plan, we explicitly minimize the embedding distance between the domains, \ie, optimizing towards better cross-domain alignment. 

Specifically, %
we convert entities (\emph{e.g.}, objects, words) in each domain (\emph{e.g.}, image, sentence) into a graph,
where each entity is represented by a feature vector, and the graph representations are recurrently updated via graph propagation. %
Cross-domain alignment can then be formulated into a graph matching problem, and be addressed by calculating matching scores based on graph distance. In our GOT framework, we utilize two types of OT distance:  
($i$) Wasserstein distance (WD)~\cite{peyre2019computational} is applied to node (entity) matching, and ($ii$) Gromov-Wasserstein distance (GWD)~\citep{peyre2016gromov} is adopted for edge (structure) matching. WD only measures the distance between node embeddings across domains, without considering topological information encoded in the graphs. GWD, on the other hand, compares graph structures 
by measuring the distance between a pair of nodes within each graph.
When fused together, the two distances allow the proposed GOT framework to effectively take into account both node and edge information for better graph matching.

The main contributions of this work are summarized as follows.
    ($i$) We propose Graph Optimal Transport (GOT), a new framework that tackles cross-domain alignment by adopting Optimal Transport for graph matching. %
    ($ii$) GOT is compatible with existing neural network models, acting as an effective drop-in regularizer to the original objective. 
    ($iii$) To demonstrate the versatile generalization ability of the proposed approach, we conduct experiments on five diverse tasks: image-text retrieval, visual question answering, image captioning, machine translation, and text summarization. Results show that GOT provides consistent performance enhancement over strong baselines across all the tasks. 

\section{Graph Optimal Transport Framework}

We first introduce the problem formulation of Cross-domain Alignment in Sec.~\ref{sec:problem_setup}, then present the proposed Graph Optimal Transport (GOT) framework in Secs.~\ref{sec:graph_contruction}-~\ref{sec:graph_matching}.

\subsection{Problem Formulation} \label{sec:problem_setup}
Assume we have two sets of entities from two different domains (denoted as $\sD_x$ and $\sD_y$). For each set, every entity is represented by a feature vector, \ie, $\tilde{\Xmat}=\{\tilde{\xv}_i\}_{i=1}^n$ and $\tilde{\Ymat}=\{\tilde{\yv}_j\}_{j=1}^m$, where $n$ and $m$ are the number of entities in each domain, respectively. The scope of this paper mainly focuses on tasks involving images and text, thus entities here correspond to objects in an image or words in a sentence. An image can be represented as a set of detected objects, each associated with a feature vector (\emph{e.g.}, from a pre-trained Faster RCNN~\cite{anderson2018bottom}). With a word embedding layer, a sentence can be represented as a sequence of word feature vectors. 

A deep neural network $f_{\thetav}(\cdot)$ can be designed to take both $\tilde{\Xmat}$ and $\tilde{\Ymat}$ as initial inputs, and generate contextualized representations:
\begin{align} \label{eqn:original_objective}
    \Xmat, \Ymat = f_{\thetav} (\tilde{\Xmat}, \tilde{\Ymat}) \,,
\end{align}
where $\Xmat=\{\xv_i\}_{i=1}^n$, $\Ymat=\{\yv_j\}_{j=1}^m$, and advanced attention mechanisms~\cite{bahdanau2014neural,vaswani2017attention} can be applied to $f_{\thetav}(\cdot)$ to simulate soft alignment. The final supervision signal $\lv$ is then used to learn $\thetav$, \emph{i.e.}, the training objective is defined as:
\begin{align}
    \Lcal(\thetav) = \Lcal_{\text{sup}}(\Xmat,\Ymat, \lv)\,.
\end{align}
Several instantiations for different tasks are summarized as follows: ($i$) \emph{Image-text Retrieval}. $\tilde{\Xmat}$ and $\tilde{\Ymat}$ are image and text features, respectively. $\lv$ is the binary label, indicating whether the input image and sentence are paired or not. Here $f_{\thetav}(\cdot)$ can be the SCAN model~\cite{SCAN}, and $\Lcal_{\text{sup}}(\cdot)$ corresponds to ranking loss~\cite{VSE,chechik2010large}. ($ii$) \emph{VQA}. Here $\lv$ denotes the ground-truth answer, $f_{\thetav}(\cdot)$ can be BUTD or BAN model~\cite{anderson2018bottom,kim2018bilinear}, $\Lcal_{\text{sup}}(\cdot)$ is cross-entropy loss. ($iii$)
\emph{Machine Translation}. $\tilde{\Xmat}$ and $\tilde{\Ymat}$ are textual features from the source and target sentences, respectively. Here $f_{\thetav}(\cdot)$ can be an encoder-decoder Transformer model~\cite{vaswani2017attention}, and $\Lcal_{\text{sup}}(\cdot)$ corresponds to cross-entropy loss that models the conditional distribution of $p(\Ymat|\Xmat)$, and here $\lv$ is not needed. To simplify subsequent discussions, all the tasks are abstracted into $f_{\thetav}(\cdot)$ and $\Lcal_{\text{sup}}(\cdot)$.  

In most previous work, the learned attention can be interpreted as a soft alignment between $\tilde{\Xmat}$ and $\tilde{\Ymat}$. However, only the final supervision signal $\Lcal_{\text{sup}}(\cdot)$ is used for model training, thus lacking an objective \emph{explicitly} encouraging cross-domain alignment. To enforce alignment and cast a regularizing effect on model training, 
we propose a new objective for Cross-domain Alignment:
\begin{align} \label{eqn:proposed_objective}
    \Lcal(\thetav) = \Lcal_{\text{sup}}(\Xmat,\Ymat, \lv) + \alpha \cdot \Lcal_{\text{CDA}} (\Xmat, \Ymat)  \,,
\end{align}
where $\Lcal_{\text{CDA}}(\cdot)$ is a regularization term that encourages alignments \emph{explicitly}, and $\alpha$ is a hyper-parameter that balances the two terms. Through gradient back-propagation, the learned $\thetav$ supports more effective relational inference.
In Section \ref{sec:graph_matching}
we describe $\Lcal_{\text{CDA}}(\cdot)$ in detail.

\subsection{Dynamic Graph Construction} \label{sec:graph_contruction}

Image and text data inherently contain rich sequential/spatial structures. By representing them as graphs and performing graph alignment, not only cross-domain relations can be modeled, but also intra-domain relations are exploited (\emph{e.g.}, semantic/spatial relations among detected objects in an image~\cite{li2019relation}). 

\begin{figure}
    \centering
    \includegraphics[width=0.46\textwidth]{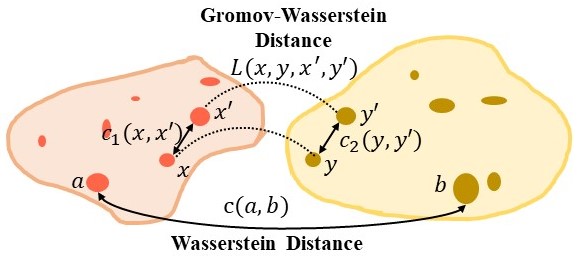}
    \caption{Illustration of the Wasserstein Distance (WD)  and the Gromov-Wasserstein Distance (GWD) used for node and structure matching, respectively. WD: $c(\av,\bv)$ is calculated between node $\av$ and $\bv$ across two domains; GWD: $L(\xv,\yv,\xv',\yv')$ is calculated between edge $c_1(\xv,\xv')$ and $c_2(\yv,\yv')$. See Sec.~\ref{sec:ot_distances} for details.}
    \label{fig:rwd}
\end{figure}

Given $\Xmat$, we aim to construct a graph $\gG_x(\sV_x, \mathcal{E}_x)$, where each node $i\in \sV_x$ is represented by a feature vector $\xv_i$.
To add edges $\mathcal{E}_x$, we first calculate the similarity between a pair of entities inside a graph: 
$\Cmat_x = \{\cos(\xv_i, \xv_j)\}_{i,j}\in \sR^{n\times n}$.
Further, we define $\Cmat_x = \max(\Cmat_x - \tau,0)$, where $\tau$ is a threshold hyper-parameter for the graph cost matrix. 
Empirically, $\tau$ is set to $0.1$. If $[\Cmat_x]_{ij}>0$, an edge is added between node $i$ and $j$. Given $\Ymat$, another graph $\gG_y(\sV_y, \mathcal{E}_y)$ can be similarly constructed.
Since both $\Xmat$ and $\Ymat$ are evolving through the update of parameters $\thetav$ during training, this graph construction process is considered ``\emph{dynamic}''.
By representing the entities in both domains as graphs, cross-domain alignment is naturally formulated into a graph matching problem.

In our proposed framework, we use Optimal Transport (OT) for graph matching, where a transport plan $\Tmat \in \R^{n\times m}$ is learned to optimize the alignment between $\Xmat$ and $\Ymat$. OT possesses several idiosyncratic characteristics that make it a good choice for solving CDA problem.
($i$) \emph{Self-normalization}: all the elements of $\Tmat^*$ sum to 1~\cite{peyre2019computational}. ($ii$) \emph{Sparsity}: when solved exactly, OT yields a sparse solution $\Tmat^*$ containing $(2r-1)$ non-zero elements at most, where $r=\max(n,m)$, leading to a more interpretable and robust alignment~\cite{de2011optimal}. ($iii$) \emph{Efficiency}: compared with conventional linear programming solvers, our solution can be readily obtained using iterative procedures that only require matrix-vector products~\cite{xie2018fast}, hence readily applicable to large deep neural networks.   

\begin{algorithm}[!t]
\caption{Computing Wasserstein Distance.}
\label{alg:wd}
\begin{algorithmic}[1]
\STATE {\bfseries Input:} \footnotesize{ $\{\xv_i\}_{i=1}^n$,$\{\yv_j\}_{j=1}^n$, $\beta$}
\STATE $\boldsymbol{\sigma}=\frac{1}{n}\mathbf{1_n}$, $\Tmat^{(1)} = \mathbf{1} \mathbf{1}^\top$
\STATE $\Cmat_{ij} = c(\xv_i, \yv_j)$, $\Amat_{ij} = {\rm e}^{-\frac{\Cmat_{ij}}{\beta}}$
\FOR{$t=1,2,3\ldots$}
    \STATE $\Qmat = \Amat \odot \Tmat^{(t)}$ \footnotesize{// $\odot$ is Hadamard product}
    \FOR{$k=1,2,3,\ldots K$}
        \STATE $\boldsymbol{\delta} = \frac{1}{n\Qmat{\boldsymbol{\sigma}}}$, $\boldsymbol{\sigma} = \frac{1}{n\Qmat^\top\boldsymbol{\delta}}$
    \ENDFOR
    \STATE $\Tmat^{(t+1)} = \text{diag}(\boldsymbol{\delta})\Qmat\text{diag}(\boldsymbol{\sigma})$
\ENDFOR
\STATE $\mathcal{D}_{wd}=\langle \Cmat^{\top}, \Tmat\rangle$ 
\STATE Return $\Tmat$, $\mathcal{D}_{w}$
// \footnotesize{$\langle \cdot, \cdot \rangle$ is the Frobenius dot-product} 
\end{algorithmic}
\end{algorithm} 
\begin{algorithm}[!t]
\caption{Computing Gromov-Wasserstein Distance.}
\label{alg:gwd}
\begin{algorithmic}[1]
\STATE {\bfseries Input:} \footnotesize{ $\{\xv_i\}_{i=1}^n$,$\{\yv_j\}_{j=1}^n$, probability vectors $\pv$, $\qv$} 

\STATE Compute intra-domain similarities:
\STATE
\quad $[\Cmat_{x}]_{ij} = \cos(\xv_i, \xv_j)$, $[\Cmat_{y}]_{ij} = \cos(\yv_i, \yv_j)$,
\STATE Compute cross-domain similarities:
\STATE \quad $\Cmat_{xy} = \Cmat_x^2 \pv \mathbf{1_m}^\top + \Cmat_y \qv (\Cmat_y^2)^\top$
\FOR{$t=1,2,3\ldots$}
    \STATE // Compute the pseudo-cost matrix
    \STATE $\Lcal = \Cmat_{xy} - 2\Cmat_x \Tmat \Cmat_y^\top$
    \STATE Apply Algorithm \ref{alg:wd} to solve transport plan $\Tmat$ 
\ENDFOR
\STATE $\mathcal{D}_{gw}=\langle \Lcal^{\top}, \Tmat\rangle$ \STATE Return $\Tmat$, $\mathcal{D}_{gw}$
\end{algorithmic}
\end{algorithm} 
\begin{figure*}[t!]
    \centering
    \includegraphics[width=1.95\columnwidth]{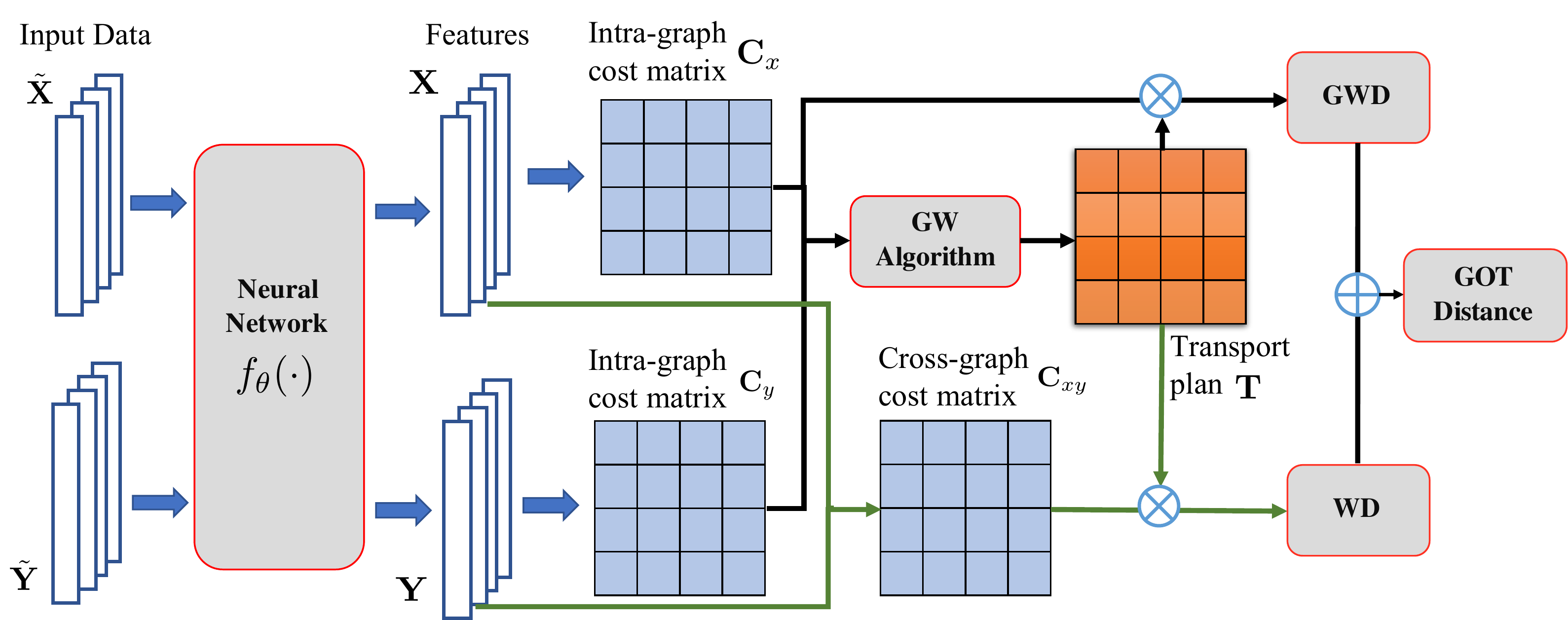}
    \caption{\label{fig:framework} Schematic computation graph of the Graph Optimal Transport (GOT) distance used for cross-domain alignment. WD is short for Wasserstein Distance, and GWD is short for Gromov-Wasserstein Distance. See Sec.~\ref{sec:problem_setup} and \ref{sec:graph_matching} for details.}
\end{figure*}

\subsection{Optimal Transport Distances} \label{sec:ot_distances}

As illustrated in Figure \ref{fig:rwd}, two types of OT distance are adopted for our graph matching: Wasserstein distance for node matching, and Gromov-Wasserstein distance for edge matching.

\paragraph{Wasserstein Distance} %
Wasserstein distance (WD) is commonly used for matching two distributions (\emph{e.g.}, two sets of node embeddings). In our setting, discrete WD can be used as a solver for network flow and bipartite matching~\citep{luise2018differential}. 
The definition of WD is described as follows.
\begin{definition}
	\label{def:wd}
	Let $\muv \in \rmP(\sX), \nuv \in \rmP(\sY)$ denote two discrete distributions, formulated as $\muv = \sum_{i=1}^n \ervu_i \delta_{\xv_i}$ and $\nuv = \sum_{j=1}^m \ervv_j \delta_{\yv_j}$, with $\delta_{\xv}$ as the Dirac function centered on $\xv$. $\Pi(\muv,\nuv)$ denotes all the joint distributions $\gammav(\xv, \yv)$, with marginals $\muv(\xv)$ and $\nuv(\yv)$.
	The weight vectors $\rvu=\{\ervu_i\}_{i=1}^n \in \Delta_n$ and $\rvv=\{\ervv_i\}_{i=1}^m \in \Delta_m$ belong to the $n$- and $m$-dimensional simplex, respectively (\ie, $\sum_{i=1}^n \ervu_i = \sum_{j=1}^m \ervv_j = 1$), where both $\muv$ and $\nuv$ are probability distributions. 
	The Wasserstein distance between the two discrete distributions $\muv,\nuv$ is defined as:
	\begin{align}\label{eq:wd}
		\mathcal{D}_{w}(\muv,\nuv) = & \inf_{\gammav\in\Pi(\muv,\nuv)}\mathbb{E}_{(\xv,\yv)\sim\gammav}\,\, [c(\xv,\yv)] \nonumber\\
		=&\min_{\Tmat\in \Pi(\rvu,\rvv)}\sum_{i=1}^n \sum_{j=1}^m \Tmat_{ij} \cdot c(\xv_i,\yv_j)\,,
	\end{align}
	where $\Pi(\rvu,\rvv) = \{ \Tmat \in \sR_+^{n\times m} | \Tmat\mathbf{1}_m=\rvu, \Tmat^\top\mathbf{1}_n=\rvv \} $, $\mathbf{1}_n$ denotes an $n$-dimensional all-one vector, and $c(\xv_i,\yv_j)$ is the cost function evaluating the distance between $\xv_i$ and $\yv_j$. For example, the cosine distance $c(\xv_i,\yv_j)=1-\frac{\xv_i^\top\yv_j}{||\xv_i||_2 ||\yv_j||_2}$ is a popular choice. The matrix $\Tmat$ is denoted as the transport plan, 
    where $\Tmat_{ij}$ represents the amount of mass shifted from $\ervu_i$ to $\ervv_j$.
\end{definition}
$\mathcal{D}_{w}(\muv,\nuv)$ defines an optimal transport distance that measures the discrepancy between each pair of samples across the two domains. In our graph matching, this is a natural choice for node (entity) matching.

\paragraph{Gromov-Wasserstein Distance} %
Instead of directly calculating distances between two sets of nodes as in WD,  
Gromov-Wasserstein distance (GWD)~\citep{peyre2016gromov, chowdhury2019gromov} can be used to calculate distances between pairs of nodes within each domain, as well as measuring how these distances compare to those in the counterpart domain. GWD in the discrete matching setting can be formulated as follows.
\begin{definition} \label{def:gwd}
	Following the same notation as in Definition \ref{def:wd}, Gromov-Wasserstein distance between $\muv, \nuv$ is defined as:
	\begin{align}\label{eq:gwd}
	&	\mathcal{D}_{gw}(\muv,\nuv) =
		 \inf_{\gammav\in\Pi(\muv,\nuv)}\mathbb{E}_{(\xv,\yv)\sim\gammav, (\xv',\yv')\sim\gammav}\,\, [L(\xv,\yv,\xv',\yv')] \nonumber \\
		&= \min_{\hat{\Tmat}\in \Pi(\rvu,\rvv)}\sum_{i,i',j,j'} \hat{\Tmat}_{ij} \hat{\Tmat}_{i'j'} L(\xv_i,\yv_j,\xv_i',\yv_j')\,,
	\end{align}
	where $L(\cdot)$ is the cost function evaluating the intra-graph structural similarity between two pairs of nodes 
	$(\xv_i,\xv_i')$ and $(\yv_j,\yv_j')$, \ie, $ L(\xv_i,\yv_j,\xv_i',\yv_j')=\|c_1(\xv_i,\xv_i') - c_2(\yv_i,\yv_i')  \|$, where $c_i,i\in[1,2]$ 
	are functions that evaluate node similarity within the same graph (\emph{e.g.}, the cosine similarity). 
\end{definition}
Similar to WD, in the GWD setting,
$c_1(\xv_i,\xv_i') $ and $ c_2(\yv_i,\yv_i')$ (corresponding to the edges) can be viewed as two nodes in the dual graphs~\citep{van2001course}, where edges are projected into nodes. 
The learned matrix $\hat{\Tmat}$ now becomes a transport plan that helps aligning the edges in different graphs.
Note that, the same $c_1$ and $c_2$ are also used for graph construction in Sec.~\ref{sec:graph_contruction}.

\subsection{Graph Matching via OT Distances} \label{sec:graph_matching}

Though GWD is capable of capturing edge similarity between graphs, it cannot be directly applied to graph alignment, since only the similarity between $c_1(\xv_i,\xv_i')$ and $ c_2(\yv_i,\yv_i')$ is considered, without taking into account node representations. For example, the word pair (``\emph{boy}", ``\emph{girl}") has similar cosine similarity as the pair (``\emph{football}", ``\emph{basketball}"), but the semantic meanings of the two pairs are completely different, and should not be matched.

On the other hand, WD can match nodes in different graphs, but fails to capture the similarity between edges. If there are duplicated entities represented by different nodes in the same graph, WD will treat them as identical and ignore their neighboring relations. For example, given a sentence
``\emph{there is a red book on the blue desk}" paired with an image containing several desks and books in different colors, it is difficult to correctly identity which book in the image the sentence is referring to,
without understanding the relations among the objects in the image. 

To best couple WD and GWD and unify these two distances in a mutually-beneficial way, we propose a transport plan $\Tmat$ shared by both WD and GWD.
Compared with naively employing two different transport plans, we observe that this joint plan works better (see Table~\ref{tab:mt_ablation_shareT}), and faster, since we only need to solve $\Tmat$ once (instead of twice).
Intuitively, with a shared transport plan, WD and GWD can enhance each other effectively, as $\Tmat$ utilizes both node and edge information simultaneously. 
Formally, the proposed GOT distance is defined as:
\begin{align} \label{eq:gotd}
    \mathcal{D}_{got}(\muv,\nuv) = & \min_{\Tmat\in \Pi(\rvu,\rvv)}\sum_{i,i',j,j'} \Tmat_{ij} \Big(\lambda c(\xv_i,\yv_j) \nonumber \\&+ (1-\lambda) \Tmat_{i'j'} \Lcal(\xv_i,\yv_j,\xv_i',\yv_j')\Big) \,.
\end{align}
We apply the Sinkhorn algorithm~\citep{cuturi2013sinkhorn, cuturi2017computational}
to solve WD (\ref{eq:wd}) with an entropic regularizer~\citep{benamou2015iterative}:
\begin{equation}
    \min_{\Tmat\in \Pi(\rvu,\rvv)}\sum_{i=1}^n \sum_{j=1}^m \Tmat_{ij} c(\xv_i,\yv_j) + \beta H(\Tmat)\,,
\end{equation}
where $H(\Tmat)=\sum_{i,j}\Tmat_{ij}\log \Tmat_{ij}$, and $\beta$ is the hyper-parameter controlling the importance of the entropy term. 
Details are provided in Algorithm \ref{alg:wd}.
The solver for GWD can be readily developed based on Algorithm \ref{alg:wd}, where $\pv,\qv$ are defined as uniform distributions (as shown in Algorithm \ref{alg:gwd}), following \citet{alvarez2018gromov}. With the help of the Sinkhorn algorithm, GOT can be efficiently implemented in popular deep learning libraries, such as PyTorch and TensorFlow.

To obtain a unified solver for the GOT distance, we define the unified cost function as:
\begin{equation}
\label{eq:ensemble}
    L_{\text{unified}} = \lambda c(\xv,\yv) + (1-\lambda) L(\xv,\yv,\xv',\yv') \,,
\end{equation}
where $\lambda$ is the hyper-parameter for controlling the importance of different cost functions.
Instead of using projected gradient descent or conjugated gradient descent as in \citet{xu2019gromov,xu2019scalable,vayer2018optimal}, we can approximate the transport plan $\Tmat$ by adding back $L_{\text{unified}}$ in Algorithm \ref{alg:gwd}, so that Line 9 in Algorithm \ref{alg:gwd} helps solve $\Tmat$ for WD and GWD at the same time, effectively matching both nodes and edges simultaneously. 
The solver for calculating the GOT distance is illustrated in Figure \ref{fig:framework}, and the detailed algorithm is summarized in Algorithm~\ref{alg:gotd}. The calculated GOT distance is used as the cross-domain alignment loss $\Lcal_{\text{CDA}}(\Xmat,\Ymat)$ in (\ref{eqn:proposed_objective}), as a regularizer to update parameters $\thetav$. 

\begin{algorithm}[!h]
\caption{Computing GOT Distance.}
\label{alg:gotd}
\begin{algorithmic}[1]
\STATE {\bfseries Input:} \footnotesize{ $\{\xv_i\}_{i=1}^n$,$\{\yv_j\}_{j=1}^m$}, hyper-parameter $\lambda$
\STATE Compute intra-domain similarities:
\STATE
\quad $[\Cmat_x]_{ij} = \cos(\xv_i, \xv_j)$, $[\Cmat_y]_{ij} = \cos(\yv_i, \yv_j)$,
\STATE $\xv'_i=g_1(\xv_i)$, $\yv'_j=g_2(\yv_j)$ // $g_1, g_2$ denote two MLPs
\STATE Compute cross-domain similarities:
\STATE \quad $\Cmat_{ij} = \cos(\xv'_i, \yv'_j)$
\IF{ $\Tmat$ is shared:}
    \STATE Update $\Lcal$ in Algorithm 2 (Line 8) with:
    \STATE \quad $\Lcal_{\text{unified}}=\lambda \Cmat + (1-\lambda) \Lcal $
    \STATE Plug in $\Lcal_{\text{unified}}$ back to Algorithm 2 and solve new $\Tmat$
    \STATE Compute $\mathcal{D}_{got}$ 
\ELSE 
    \STATE Apply Algorithm 1 to obtain $\mathcal{D}_{w}$
    \STATE Apply Algorithm 2 to obtain $\mathcal{D}_{gw}$
    \STATE $\mathcal{D}_{got} = \lambda \mathcal{D}_{w} + (1-\lambda) \mathcal{D}_{gw}$
\ENDIF
\STATE Return $\mathcal{D}_{got}$
\end{algorithmic}
\end{algorithm} 

\section{Related Work}
\paragraph{Optimal Transport}
Wasserstein distance (WD), a.k.a. Earth Mover's distance, has been widely applied to machine learning tasks. %
In computer vision, \citet{rubner1998metric} uses WD to discover the structure of color distribution for image search.
In natural language processing, WD has been applied to document retrieval~\cite{kusner2015word} and sequence-to-sequence learning~\cite{chen2019improving}.
There are also studies adopting WD in Generative Adversarial Network (GAN)~\cite{goodfellow2014generative, salimans2018improving,chen2018adversarial,mroueh2018sobolev,zhang2020nested} to alleviate the mode-collapse issue. Recently, it has also been used for vision-and-language pre-training to encourage word-region alignment~\cite{chen2019uniter}. 
Besides WD, Gromov-Wassersten distance~\cite{peyre2016gromov} has been proposed for distributional metric matching and applied to unsupervised machine translation~\cite{alvarez2018gromov}.

There are different ways to solve the OT distance, such as linear programming. However, this solver is not differentiable, thus it cannot be applied in deep learning frameworks. Recently, WGAN~\cite{arjovsky2017wasserstein} proposes to approximate the dual form of WD by imposing a 1-Lipschitz constraint on the discriminator.
Note that the duality used for WGAN is restricted to the W-1 distance, $i.e.$, $\|\cdot\|$.
The Sinkhorn algorithm was first proposed in \citet{cuturi2013sinkhorn} as a solver for calculating an entropic regularized OT distance. Thanks to the Envelop Theorem~\citep{cuturi2017computational}, the Sinkhorn algorithm can be efficiently calculated and readily applied to neural networks.
More recently, \citet{vayer2018optimal} proposed the fused GWD for graph matching. 
Our proposed GOT framework enjoys the benefits of both Sinkhorn algorithm and fused GWD: it is ($i$) capable of capturing more structured information via marrying both WD and GWD; and ($ii$) scalable to large datasets and trainable with deep neural networks.

\begin{table*}[t!]
  \begin{center}
  \centering
  \small
    \begin{tabular*}{1.\linewidth}{@{\extracolsep{\fill}}lccccccc}
         \toprule
         & \multicolumn{3}{c}{Sentence Retrieval} & \multicolumn{3}{c}{Image Retrieval} & \multicolumn{1}{c}{}\\
        Method & R@1 & R@5 & R@10 & R@1 & R@5 & R@10 &Rsum\\
        \toprule
        VSE++ (ResNet) \cite{VSE} & 52.9 & -- & 87.2 & 39.6 & -- & 79.5 & -- \\ 
        DPC (ResNet) \cite{zheng2017dual} & 55.6 & 81.9 & 89.5 & 39.1 & 69.2 & 80.9 & 416.2\\ 
        DAN (ResNet) \cite{nam2017dual} & 55.0 & 81.8 & 89.0 & 39.4 & 69.2 & 79.1 & 413.5\\ 
        SCO (ResNet) \cite{huang2018learning} & 55.5 & 82.0 & 89.3 & 41.1 & 70.5 & 80.1 & 418.5\\ 
        SCAN (Faster R-CNN, ResNet) \cite{SCAN}  & 67.7 & 88.9 & 94.0 & 44.0 & 74.2 & 82.6 &452.2 \\ 
        \textbf{Ours (Faster R-CNN, ResNet)}:\\
        SCAN + WD  &{70.9} &{92.3} &{95.2} &{49.7} &{ 78.2} &{86.0} &{472.3
}\\
        SCAN + GWD  &{69.5} &{91.2} &{95.2} &{48.8} &{78.1} &{85.8} &{468.6}\\
        SCAN + GOT  &\textbf{70.9} &\textbf{92.8} &\textbf{95.5} &\textbf{50.7} & \textbf{78.7} &\textbf{86.2} &\textbf{474.8}\\
        \hline
        \noalign{\smallskip}
        VSE++ (ResNet) \cite{VSE} & 41.3 & -- & 81.2 & 30.3 & -- & 72.4 & -- \\ 
        DPC (ResNet) \cite{zheng2017dual} & 41.2 & 70.5 & 81.1 & 25.3 & 53.4 & 66.4 & 337.9\\ 
        GXN (ResNet) \cite{gu2018look} & 42.0 & -- & 84.7 & 31.7 & -- & 74.6 & -- \\ 
        SCO (ResNet) \cite{huang2018learning} & 42.8 & 72.3 & 83.0 & 33.1 & 62.9 & 75.5 & 369.6 \\ 
        SCAN (Faster R-CNN, ResNet)\cite{SCAN} & 46.4 & 77.4 & 87.2 & 34.4 & 63.7 & 75.7 & 384.8 \\ 
        \textbf{Ours (Faster R-CNN, ResNet)}:\\
        SCAN + WD &{50.2} &{80.1} &{89.5} &{37.9} &{66.8} &{78.1} &{402.6}\\
        SCAN + GWD &{47.2} & {78.3} &{87.5} &{34.9} &{64.4} &{76.3} &{388.6
}\\
        SCAN + GOT &\textbf{50.5} &\textbf{80.2} &\textbf{89.8} &\textbf{38.1} & \textbf{66.8} &\textbf{78.5} &\textbf{403.9}\\
        \bottomrule
    \end{tabular*}
    \vspace{-2mm}
     \caption{\label{table:retrieval} Results on image-text retrieval evaluated on Recall@$K$ (R@$K$). Upper panel: Flickr30K; lower panel: COCO. 
  }
  \end{center}
\end{table*} 
\paragraph{Graph Neural Network}
Neural networks operating on graph data was first introduced in \citet{gori2005new} using recurrent neural networks.
Later, \citet{duvenaud2015convolutional} proposed a convolutional neural network over graphs for classification tasks. However, these methods suffer from scalability issues, because they need to learn node-degree-specific weight matrices for large graphs.
To alleviate this issue, \citet{kipf2016semi} proposed to use a single weight matrix per layer in the neural network, which is capable of handling varying node degrees through an appropriate normalization of the adjacency matrix of the data.
To further improve the classification accuracy, the graph attention network (GAT) \cite{velivckovic2017graph} was proposed by using a learned weight matrix instead of the adjacency matrix, with masked attention to aggregate node neighborhood information.

Recently, the graph neural network has been extended to other tasks beyond classification. \citet{li2019graph} proposed graph matching network (GMN) for learning similarities between graphs. Similar to GAT, masked attention is applied to aggregate information from each node within a graph, and cross-graph information is further exploited via soft attention. Task-specific losses are then used to guide model training. 
In this setting, an adjacency matrix can be directly obtained from the data and 
soft attention is used to induce alignment. 
In contrast, our GOT framework does not rely on explicit graph structures in the data, and uses OT for graph alignment. %

\begin{figure*}[t!]
\centering
    \includegraphics[width=0.95\textwidth]{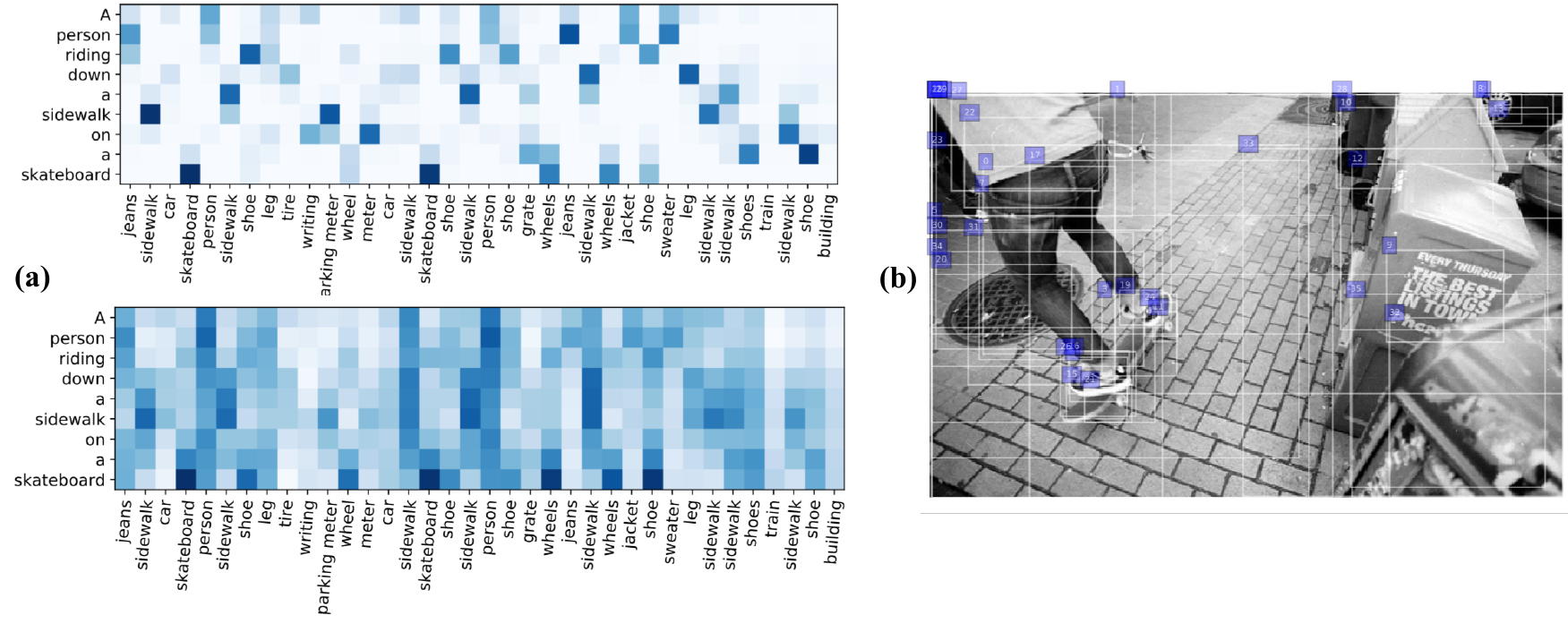}
    \vspace{-2mm}
    \caption{\label{fig:GOTvsAttention}(a) A comparison of the inferred transport plan from GOT (top chart) and the learned attention matrix from SCAN (bottom chart). Both serve as a lens to visualize cross-domain alignment. The horizontal axis represents image regions, and the vertical axis represents word tokens. (b) The original image. }
\end{figure*}

\section{Experiments}
To validate the effectiveness 
of the proposed GOT framework, we evaluate performance on a selection of diverse tasks. We first consider vision-and-language understanding, including: ($i$) image-text retrieval, and ($ii$) visual question answering. We further consider text generation tasks, including: ($iii$) image captioning, ($iv$) machine translation, and ($v$) abstractive text summarization. Code is available at \url{https://github.com/LiqunChen0606/Graph-Optimal-Transport}.

\subsection{Vision-and-Language Tasks}

\paragraph{Image-Text Retrieval}
For image-text retrieval task, we use pre-trained Faster R-CNN~\cite{ren2015faster} to extract bottom-up-attention features~\cite{anderson2018bottom} as the image representation. A set of 36 features is created for each image, each feature represented by a 2048-dimensional vector. For captions, a bi-directional GRU~\cite{schuster1997bidirectional,bahdanau2014neural} is used to obtain textual features.

We evaluate our model on the Flickr30K \cite{plummer2015flickr30k} and COCO \cite{lin2014microsoft} datasets. Flickr30K contains $31$,$000$ images, with five human-annotated captions per image. We follow previous work~\cite{karpathy2015deep,VSE} for the data split: $29$,$000$, $1$,$000$ and $1$,$000$ images are used for training, validation and test, respectively.
COCO contains $123$,$287$ images, each image also accompanied with five captions. 
We follow the data split in \citet{VSE}, where $113$,$287$, $5$,$000$ and $5$,$000$ images are used for training, validation and test, respectively.

We measure the performance of image retrieval and sentence retrieval on Recall at $K$ (R@$K$)~\cite{karpathy2015deep}, defined as the percentage of queries retrieving the correct images/sentences within the top $K$ highest-ranked results. In our experiment, $K=\{1,5,10\}$, and Rsum~\cite{huang2017instance} (summation over all R@$K$) 
is used to evaluate the overall performance. Results are summarized in Table \ref{table:retrieval}. Both WD and GWD can boost the performance of the SCAN model, while WD achieves a larger margin than GWD. This indicates that when used alone, GWD may not be a good metric for graph alignment.
When combining the two distances together, GOT achieves the best performance. 

Figure \ref{fig:GOTvsAttention} provides visualization on the learned transport plan in GOT and the learned attention matrix in SCAN. Both serve as a proxy to lend insights into the learned alignment. As shown, the attention matrix from SCAN is much denser and noisier than the transport plan inferred by GOT. 
This shows our model can better discover cross-domain relations between image-text pairs, since the inferred transport plan is more interpretable and has less ambiguity. For example, both the words ``\emph{sidewalk}'' and ``\emph{skateboard}'' match the corresponding image regions very well.

Because of the Envelope Theorem~\cite{cuturi2017computational}, GOT needs to be calculated only during the forward phase of model training. Therefore, it does not introduce much extra computation time.
For example, when using the same machine for image-text retrieval experiments, SCAN required 6hr 34min for training and
SCAN+GOT 6hr 57min.

\paragraph{Visual Question Answering}

\begin{table}[!t]
    \centering
    \small
    \begin{tabular}{l c c c c} %
    \hline
    {\bf Model}  &  {\bf BAN} & {\bf BAN+GWD} & {\bf BAN+WD} & {\bf BAN+GOT}\\ 
    \hline
    
    {\bf Score} & 66.00 & 66.21 & 66.26 & \textbf{66.44}\\

    \hline
    \end{tabular}
    \vspace{-3mm}
    \caption{\label{tab:vqa} Results (accuracy) on VQA 2.0 validation set, using BAN~\cite{kim2018bilinear} as baseline.}
\end{table}
\begin{table}[!t]
    \centering
    \small
    \resizebox{0.48\textwidth}{!}{
    \begin{tabular}{l c c c c c} %
    \hline
    {\bf Model}  & {\bf BUTD} & {\bf BAN-1} & {\bf BAN-2} & {\bf BAN-4} & {\bf BAN-8}\\ 
    \hline
    {\bf w/o GOT }  & 63.37 & 65.37 & 65.61 & 65.81 & 66.00 \\
    {\bf w/ GOT } & \textbf{65.01} & \textbf{65.68} & \textbf{65.88} & \textbf{66.10} & \textbf{66.44} \\
    \hline
    \end{tabular}
    }
    \vspace{-3mm}
    \caption{\label{tab:vqa_dfBAN} Results (accuracy) of applying GOT to BUTD~\cite{anderson2018bottom} and BAN-$m$~\cite{kim2018bilinear} on VQA 2.0. $m$ denotes the number of glimpses.}
\end{table} 
We also consider the VQA 2.0 dataset~\cite{goyal2017making}, 
which contains human-annotated QA pairs on COCO images~\citep{lin2014microsoft}. For each image, an average of $3$ questions are collected, with $10$ candidate answers per question. The most frequent answer from the annotators is selected as the correct answer. Following previous work~\cite{kim2018bilinear}, we take the answers that appear more than 9 times in the training set as candidate answers, which results in $3129$ candidates. Classification accuracy is used as the evaluation metric, defined as $\min (1, \frac{\text{\# humans provided ans.}}{3})$.

The BAN model~\cite{kim2018bilinear} is used as baseline, with the original codebase used for fair comparison. Results are summarized in Table \ref{tab:vqa}. Both WD and GWD improve the BAN model on the validation set, and GOT achieves further performance lift.

We also investigate whether different architecture designs affect the performance gain. We consider BUTD~\cite{anderson2018bottom} as an additional baseline, and apply different number of glimpses $m$ to the BAN model, denoted as BAN-$m$. Results are summarized in Table~\ref{tab:vqa_dfBAN},
with the following observations: ($i$) When the number of parameters in the tested model is small, such as BUTD, the improvement brought by GOT is more significant.
($ii$) BAN-4, a simpler model than BAN-8, when combined with GOT, can outperform BAN-8 without using GOT (66.10 v.s. 66.00).
($iii$) For complex models such as BAN-8 that might have limited space for improvement, GOT is still able to achieve performance gain. %
\begin{table*}[ht!]\small
    \centering
\resizebox{0.92\textwidth}{!}{
\begin{tabular}{lccccccc}
\hline
Method           & CIDEr & BLEU-4 & BLUE-3 & BLEU-2 & BLEU-1 & ROUGE & METEOR \\ \hline
Soft Attention~\citep{xu2015show}  & - & 24.3 & 34.4 & 49.2 & 70.7 & - &23.9\\
Hard Attention~\citep{xu2015show}  &  - & 25.0 & 35.7 & 50.4 & 71.8 & -&23.0 \\
Show \& Tell~\citep{vinyals2015show} & 85.5 & 27.7    & -    & - & - & -& 23.7 \\
ATT-FCN~\citep{you2016image} &    - & 30.4 & 40.2 & 53.7 & 70.9 & - & 24.3\\
SCN-LSTM~\citep{gan2017semantic} & 101.2 & 33.0 & 43.3 & 56.6 & 72.8 & -& 25.7\\
Adaptive Attention~\citep{lu2017knowing}  &  108.5 & 33.2 & 43.9 &  58.0 &  74.2 & -& 26.6 \\
\hline
MLE              & 106.3 & 34.3   & 45.3   & 59.3   & 75.6   & 55.2  & 26.2   \\
MLE + WD        & 107.9 & {34.8}   & 46.1   & 60.1   & 76.2   & 55.6  & 26.5   \\
MLE + GWD          & 106.6 & {33.3}   & 45.2   & 59.1   & 75.7   & 55.0  & 25.9   \\
MLE + GOT  & \textbf{109.2} & \textbf{35.1}   & \textbf{46.5}   & \textbf{60.3}   & \textbf{77.0}   & \textbf{56.2}  & \textbf{26.7}   \\ \hline
\end{tabular}}
\caption{ \label{table:caption}Results of image captioning on the COCO dataset. }
\end{table*} %

\begin{table*}[!t]
    \centering
    \small
    \begin{tabular}{l c c c c} %
    \hline
    {\bf Model}  & {\bf EN-VI uncased} & {\bf EN-VI cased} & {\bf EN-DE uncased} & {\bf EN-DE cased}\\ 
    \hline
    Transformer~\cite{vaswani2017attention}    & $29.25 \pm 0.18$ & $28.46 \pm 0.17$ & $25.60 \pm 0.07$ & $25.12 \pm 0.12$\\
    Transformer + WD  & $29.49 \pm 0.10$ & $28.68 \pm 0.14$ & $25.83 \pm 0.12$ & $25.30 \pm 0.11$\\
    Transformer + GWD & $28.65 \pm 0.14$ & $28.34 \pm 0.16$ & $25.42 \pm 0.17$ & $24.82 \pm 0.15$\\
    Transformer + GOT & $\mathbf{29.92 \pm 0.11}$ & $\mathbf{29.09 \pm 0.18}$ & $\mathbf{26.05 \pm 0.17}$ & $\mathbf{25.54 \pm 0.15}$\\
    \hline
    \end{tabular}
    \vspace{-2mm}
    \caption{\label{tab:mt} Results of neural machine translation on EN-DE and EN-VI.}
\end{table*}
\begin{table}[t!]
    \centering
    \small
    \resizebox{0.48\textwidth}{!}{
    \begin{tabular}{l c c c} %
    \hline
    {\bf Method} & {\bf ROUGE-1} & {\bf ROUGE-2} & {\bf ROUGE-L}\\ 
    \hline
    ABS+~\cite{rush2015neural} & $31.00$  & $12.65$ & $28.34$ \\
    LSTM~\cite{hu2018texar} & $36.11 $ & $16.39 $ & $32.32 $ \\
    LSTM + GWD  & $36.31 $ & $17.32 $ & $33.15 $ \\
    LSTM + WD  & $36.81 $ & $17.34 $ & $33.34 $ \\
    LSTM + GOT & \bm{$37.10$} & \bm{$17.61 $} & \bm{$33.70 $} \\
    \hline
    \end{tabular}
    }
    \vspace{-2mm}
    \caption{\label{tab:giga} Results of abstractive text summarization on the English Gigawords dataset.}
\end{table}

\subsection{Text Generation Tasks}
\paragraph{Image Captioning}
We conduct experiments on image captioning using the same COCO dataset. The same bottom-up-attention features~\cite{anderson2018bottom} used in image-text retrieval are adopted here. The text decoder is one-layer LSTM with 256 hidden units. The word embedding dimension is set to 256. 
Results are summarized in Table \ref{table:caption}.
A similar performance gain is introduced by GOT. The relative performance boost from WD to GOT over CIDEr score is: $\frac{\text{GOT}-\text{WD}}{\text{WD}-\text{MLE}}=\frac{109.2-107.9}{107.9-106.3}=81.25\%$. 
This attributes to the additional GWD introduced in GOT that can help model implicit intra-domain relationships in images and captions, leading to more accurate caption generation. 

\paragraph{Machine Translation}
In machine translation (and abstractive summarization), the word embedding spaces of the source and target sentences are different, which can be considered as different domains. Therefore, GOT can be used to align those words with similar semantic meanings between the source and target sentences for better translation/summarization.
We choose two machine translation benchmarks for experiments: ($i$) English-Vietnamese TED-talks corpus, which contains $133$K pairs of sentences from the IWSLT Evaluation Campaign~\citep{cettolo2015iwslt}; and ($ii$) a large-scale English-German parallel corpus with $4.5$M pairs of sentences, from the WMT Evaluation Campaign~\citep{vaswani2017attention}. 
The Texar codebase~\citep{hu2018texar} is used in our experiments. 

We apply GOT to the Transformer model~\citep{vaswani2017attention} and use BLEU score~\cite{papineni2002bleu} as the evaluation metric.
Results are summarized in Table \ref{tab:mt}.
As also observed in \citet{chen2019improving}, using WD can improve the performance of the Transformer for sequence-to-sequence learning. 
However, if only GWD is used, the test BLEU score drops. Since GWD can only match
the edges, it ignores supervision signals from node representations. This serves as empirical evidence to support our hypothesis that using GWD alone may not be enough to improve performance. 
However, GWD serves as a complementary method for capturing graph information that might be missed by WD. Therefore, when combining the two together, GOT achieves the best performance. Example translations are provided in Table~\ref{tab:translation_example}.

\begin{figure}
    \centering
    \includegraphics[width=0.46\textwidth]{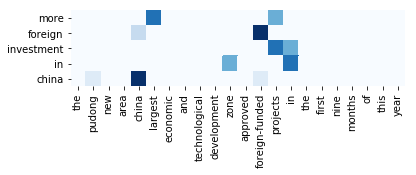}
    \vspace{-2mm}
    \caption{Inferred transport plan for aligning source and output sentences in abstractive summarization.}
    \label{fig:abs_plan}
    \vspace{-3mm}
\end{figure}

\begin{table*} [t!] 
    \small
	\centering
	\begin{tabular}{c l}
        \vspace{1mm} \\ 
        \hline  \vspace{0mm}  \\
        {\bf Reference}: & India’s new prime minister, Narendra Modi, is meeting his Japanese counterpart, Shinzo Abe, in Tokyo to discuss  \\ & economic and 
        security ties, on his first major foreign visit \textcolor{blue}{since} \textcolor{red}{winning} \textcolor{blue}{May’s election}. 
         \vspace{1mm} \\ 
        {\bf MLE}: & India ‘ s new prime minister , Narendra Modi , meets his Japanese counterpart , Shinzo Abe , in Tokyo , during his  \\ 
        & first major foreign visit \textcolor{blue}{in May} to discuss economic and security relations . \vspace{1mm} \\
        {\bf GOT}: & India ’ s new prime minister , Narendra Modi , is meeting his Japanese counterpart Shinzo Abe in Tokyo in his first \\ & major foreign visit \textcolor{blue}{since his election} \textcolor{red}{victory} \textcolor{blue}{in May} to discuss economic and security relations.
    \vspace{1mm} \\ 
        \hline  \vspace{0mm}  \\
        {\bf Reference}: & Chinese leaders presented the Sunday ruling as a democratic breakthrough because it gives Hong Kongers a direct \\ 
        & vote, but the decision also  makes clear that Chinese leaders would retain a firm hold on the process through a  \\ 
        & nominating committee tightly controlled by \textcolor{blue}{Beijing}. \vspace{1mm} \\ 
        {\bf MLE}: & The Chinese leadership presented the decision of Sunday as a democratic breakthrough , because it gives Hong  \\ 
        & Kong citizens a direct  right to vote , but the decision also makes it clear that the Chinese leadership maintains the  \\ 
        & \textbf{expiration of} a nomination committee closely controlled  by \textcolor{blue}{Beijing} . \vspace{1mm} \\
        
        {\bf GOT}: & The Chinese leadership presented the decision on Sunday as a democratic breakthrough , because Hong Kong  \\ 
        & citizens have a direct electoral right , but the decision also makes it clear that the Chinese leadership remains  \\ 
        & firmly in hand with a nominating committee controlled  by \textcolor{blue}{Beijing}. \vspace{1mm} \\
        \hline
	\end{tabular}

	\caption{Comparison of German-to-English translation examples. For each example, we show the human translation (reference) and the translation from MLE and GOT. We highlight the key-phrase differences between reference and translation outputs in blue and red, and denote the error in translation in bold.  In the first example, GOT correctly maintains all the information in {\em ``since winning May's election"} by translating to {\em ``since his election victory in May"}, whereas MLE only generate {\em ``in May"}. In the second example, GOT successfully keeps the information {\em ``Beijing"}, whereas MLE generates wrong words {\em ``expiration of"}.}
	\label{tab:translation_example}
\end{table*}

\paragraph{Abstractive Summarization} %
We evaluate abstractive summarization on the English Gigawords benchmark~\citep{graff2003english}.
A basic LSTM model as implemented in Texar~\citep{hu2018texar} is used in our experiments. ROUGE-1, -2 and -L scores~\cite{lin2004rouge} are reported. 
Table \ref{tab:giga} shows that both GWD and WD can improve the performance of the LSTM. 
The transport plan for source and output sentences alignment is illustrated in Figure \ref{fig:abs_plan}. The learned alignment is sparse and interpretable. For instance, the words ``\emph{largest}'' and ``\emph{projects}'' in the source sentence matches the words ``\emph{more}'' and ``\emph{investment}'' in the output summary very well.  

\subsection{Ablation study}
We conduct additional ablation study on the EN-VI and EN-DE datasets for machine translation.
\paragraph{Shared Transport Plan $\Tmat$}
As discussed in Sec. \ref{sec:graph_matching}, we use a shared transport plan $\Tmat$ to solve the GOT distance. An alternative is not to share this $\Tmat$ matrix. 
The comparison results are provided in Table \ref{tab:mt_ablation_shareT}. GOT with a shared transport plan achieves better performance than the alternative. Since we only need to run the iterative Sinkhorn algorithm once, it also saves training time than the unshared case.

\begin{table}[t]
    \centering
    \small
    \begin{tabular}{l c c } %
    \hline
    {\bf Model}  & {\bf EN-VI uncased} & {\bf EN-DE uncased} \\ 
    \hline
    GOT (shared) & ${29.92 \pm 0.11}$ & ${26.05 \pm 0.18}$  \\
    GOT (unshared) & ${29.77 \pm 0.12}$ & ${25.89 \pm 0.17}$  \\
    \hline
    \end{tabular}
    \caption{\label{tab:mt_ablation_shareT} Ablation study on transport plan in machine translation. Both models were run 5 times with the same hyper-parameter setting.}
\end{table}

\begin{table}[!t]
    \centering
    \small
    \resizebox{0.48\textwidth}{!}{
    \begin{tabular}{l c  c c c c c} %
    \hline
    {\bf $\lambda$}  & 0  & 0.1 & 0.3 & 0.5 & 0.8 & 1.0\\ 
    \hline
    
    {\bf BLEU} & $28.65$  & $29.31$ & $29.52$ & $29.65$ & {\bf 29.92} & $29.49$\\

    \hline
    \end{tabular}
    }
    \vspace{-3mm}
    \caption{\label{tab:lambda_ab} Ablation study of the hyper-parameter $\lambda$ on the EN-VI machine translation dataset.}
    \vspace{-3mm}
\end{table}
 
\paragraph{Hyper-parameter $\lambda$} We perform ablation study on the hyper-parameter $\lambda$ in (\ref{eq:gotd}). We select
$\lambda$ from $[0,1]$ and report results in Table \ref{tab:lambda_ab}. When $\lambda=0.8$, EN-VI translation performs the best, which indicates that the weight on WD needs to be larger than the weight on GWD, since intuitively node matching is more important than edge matching for machine translation. 
However, both WD and GWD contribute to GOT achieving the best performance.

\section{Conclusions}
We propose Graph Optimal Transport, a principled framework for cross-domain alignment. With the Wasserstein and Gromov-Wasserstein distances, both intra-domain and cross-domain relations are captured for better alignment. Empirically, we observe that enforcing alignment can serve as an effective regularizer for model training. Extensive experiments show that the proposed method is a generic framework that can be applied to 
a wide range of cross-domain tasks.
For future work, we plan to apply the proposed framework to self-supervised representation learning.

\section*{Acknowledgements}

The authors would like to
thank the anonymous reviewers for their insightful
comments. The research at Duke University was supported in part by
DARPA, DOE, NIH, NSF and ONR.

\bibliography{icml_refs}
\bibliographystyle{icml2020}

\end{document}